RESEARCH

# A portable solution for simultaneous human movement and mobile EEG acquisition: readiness potential for basketball free-throw shooting

**Miguel Contreras-Altamirano[1] · Melanie Klapprott[1] · Nadine Jacobsen[1] · Paul Maanen[1,2] · Julius Welzel[5] · Stefan Debener[1,2,3,4]**

## Abstract

Advances in wireless electroencephalography (EEG) technology promise to record brain-electrical activity in everyday situations. To better understand the relationship between brain activity and natural behavior, it is necessary to monitor human movement patterns. Here, we present a pocketable setup consisting of two smartphones to simultaneously capture human posture and EEG signals. We asked 26 basketball players to shoot 120 free throws each. First, we investigated whether our setup allows us to capture the readiness potential (RP) that precedes voluntary actions. Second, we investigated whether the RP differs between successful and unsuccessful free-throw attempts. The results confirmed the presence of the RP over fronto-central channels, with significant negative deflection at channel Cz, from −400 to 0 ms before movement onset ($M \pm SE$: − 6.54 ± 2.26 to − 13.52 ± 2.42 µV; $z = -2.53$ to − 3.92; FDR-corrected $p = 0.049$ to 0.003; $r = 0.50$ to 0.77). However, the amplitude of the RP was not related to shooting success (all FDR-corrected $p > 0.05$; maximum mean $R2 = 0.047$, i.e., 4.7% explained variance). Preliminary exploratory pose analysis conducted offline indicated the presence of participant-specific variations in posture between successful and unsuccessful shots in 38.5% of participants (10/26), with 4.5% explained variance (maximum mean landmark $R2 = 0.045$). We conclude that a highly portable, low-cost and lightweight acquisition setup, consisting of two smartphones and a head-mounted wireless EEG amplifier, is sufficient to monitor complex human movement patterns and associated brain dynamics outside the laboratory.

## Introduction

Technical limitations typically restrict non-invasive functional brain research to stationary, controlled recording conditions. Methods such as functional magnetic resonance imaging (fMRI) and magnetoencephalography (MEG) are particularly sensitive to head movement, and electroencephalography (EEG) recordings are also susceptible to movement-related artifacts. Consequently, much of what we know about the brain is therefore based on simplistic and unnatural stationary recording settings (Brunswik 1956; Nastase et al. 2020; Shamay-Tsoory and Mendelsohn 2019). Accordingly, the results may not generalize well to natural, active and mobile situations. This raises concerns about ecological validity (EV), which assesses whether measurements and behaviors in research settings are representative of the real world (Chang et al. 2022). Technological developments, such as advancements in wireless amplifiers, lightweight recording hardware, synchronization techniques and signal processing methods, have increasingly enabled EEG and functional near-infrared spectroscopy (fNIRS) to be adapted for mobile and real-world applications. This makes it easier to study the brain outside the laboratory (Boere et al. 2024; Debener et al. 2012, 2015; Bleichner and Debener 2017; Gramann et al. 2011; Moffat et al., 2023). The usability of mobile EEG has been demonstrated in various whole-body movement scenarios, such as walking (Debener et al. 2012; De Vos et al. 2014; Jacobsen et al. 2022; Straetmans et al. 2022, 2024), cycling (Scanlon et al. 2019; Zink et al. 2016), skateboarding (Robles et al. 2021; Callan et al. 2024), slacklining (Papin et al. 2024), and even freestyle swimming (Klapprott and Debener 2024). These studies were able to record brain-electrical activity reflecting sensory, cognitive and motor functions. As these processes also play an important role in athletic performance (Baumeister et al. 2008; Bertollo et al. 2016; Cheron et al. 2016; Crews and Landers 1993; Hunt et al. 2013; Tan et al. 2019), several groups have started to investigate the neural dynamics underlying athletic performance using mobile EEG (Park et al. 2015; Schäring et al. 2023).

✉ Miguel Contreras-Altamirano
miguel.angel.contreras.altamirano@uni-oldenburg.de

[1] Neuropsychology Lab, Department of Psychology, School of Medicine and Health Sciences, Carl Von Ossietzky Universität Oldenburg, Oldenburg, Germany

[2] Cluster of Excellence "Hearing4All", Carl Von Ossietzky Universität Oldenburg, Oldenburg, Germany

[3] Fraunhofer Institute of Digital Media Technology, Oldenburg Branch for Hearing, Oldenburg, Germany

[4] Center for Neurosensory Science and Systems, Carl Von Ossietzky University of Oldenburg, Oldenburg, Germany

[5] Kiel University, Kiel, Germany

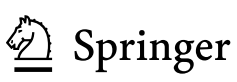

Complex motor skills play a significant role in the life course of individuals (Espenhahn et al. 2019; Etnier et al. 1996; Veldman et al. 2018). Whereas closed skill sports involve actions directed toward stable and predictable stimuli, open skill sports involve activities directed toward stimuli in unpredictable and dynamic environments (Gu et al. 2019). Exercise neuroscience research categorizes basketball as involving goal-directed skills, including target-specific motor skills, because it requires the execution of precise movements directed toward a specific target location, namely the basket. Evidently, throwing a basketball toward a target requires a highly precise sequence of movements (Cheng et al. 2015a, b, Chang et al. 2022, Feng et al. 2023, O'Brien et al. 2017, Qiu et al. 2019, Shi and Feng 2022). To date, studies have analyzed brain function during free-throw shooting using simulated basketball scenarios or stationary conditions (Chien-Ting et al., 2007; Chuang et al. 2013; Keshvari et al. 2023; Kanatschnig et al. 2023; Moscaleski et al. 2022; Ramezanzade et al. 2023; Robin et al. 2019).

EEG can capture several neural markers relevant to movement, including oscillatory dynamics (e.g. gait-cycle-related modulations of alpha- and beta-band power; Gwin et al. 2011), event-related potentials (Debener et al. 2012), and movement-related cortical potentials (Shibasaki and Hallett 2006). Among these markers, the readiness potential (RP) is particularly relevant for the preparatory phase of voluntary movement and can therefore be analyzed before large movement artifacts dominate the EEG signal. The RP is a slow negative shift in brain-electrical activity that precedes voluntary, self-initiated movements and reflects cortical brain activity involved in the preparation of a motor action (Kornhuber and Deecke 1965). Functionally, the RP originates in the frontal cortex and has been associated with skill acquisition and decision making (Lui et al. 2021; Mann et al. 2011; Nann et al. 2019; Vogt et al. 2017). Studies relating the RP to target motor skills have found greater RP amplitude in professionals compared to amateurs (Mann et al. 2011; Vogt et al. 2017). As the RP occurs during the preparation or planning phase of a movement, it might be expected to be related to movement efficiency and performance. However, most RP studies have not investigated complex human behavior, but rather isolated finger button presses (Schurger et al. 2021). One exception is a study that identified the RP before bungee jumping and jumping from 1-m height, but motor performance was not assessed (Nann et al. 2019). No previous report compared the RP for successful and unsuccessful basketball shots. Other studies have evaluated basketball shooting performance using motion sensors (Kuhlman and Min 2021; Lian et al. 2021; Shankar et al. 2018), body position capture (Ji 2020), ball trajectory analysis (Zhao et al. 2022), and personality traits (Siemon and Wessels 2023), partially under laboratory conditions.

The concept of mobile brain/body imaging (MoBI) aims to study human brain activity and motor behavior in naturalistic environments (Gramann et al. 2011). While the simultaneous recording of brain-electrical activity and complex whole-body motion patterns promises a more comprehensive understanding of natural human behavior (Debener et al. 2012; Ladouce et al. 2016; Makeig et al. 2009; Jungnickel et al. 2019), many studies in this area typically use virtual reality and stationary motion capture technologies, which may however compromise the mobility of the device and participants (Bateson et al. 2017). These considerations highlight the need for integrated brain–body measurements in which neural activity is recorded together with concurrent movement signals during natural behavior. Here, we report a novel, portable and low-cost approach to simultaneously record EEG and human whole-body actions on smartphones outside of the laboratory. We examined cortical motor preparation for basketball free-throws by combining wireless EEG, an offline artificial intelligence (AI) based real-time pose detection, and a wrist-worn inertial measurement unit (IMU). First, we investigated whether the setup enables the detection of the RP prior to movement onset during basketball free-throws. Second, we tested whether the RP or human pose markers could differentiate between successful (hits) and unsuccessful (misses) basketball shots. This proof-of-principle study aims to demonstrate the feasibility of combining mobile EEG and smartphone-based human pose tracking for future brain–body research in naturalistic settings.

## Materials and methods

### Participants

26 right-handed, healthy participants (3 female, 23 male) participated in this study. Participants performed basketball free-throws while their brain activity and movement patterns were recorded. Participants' ages ranged from 18 to 32 years ($M$ = 25.88 years, $SD$ = 4.19 years). They were recruited via the bulletin board of the online campus system of the Oldenburg University, mailing lists, personal contacts, social media, newspapers as well as by contacting regional sports clubs. They all reported having at least 3 years of experience as basketball players and played the sport regularly (at least twice per week). All participants reported normal or corrected to normal vision. The specifications of the basketball profile for all participants can be found in Supplementary Table 1. Exclusion criteria were neurological disease, drug use, alcohol consumption prior to the day of the examination, and use of any medication. Volunteers provided their written informed consent prior to participation, and were compensated with 10€/h. The experiment was conducted according to the tenets of the 1964 Declaration of Helsinki and with the approval of the ethics committee of the University of Oldenburg (approval number: Drs.EK/2023/087).

In the absence of comparable prior research, an approximate estimate based on a statistical power calculation was performed to determine the appropriate sample size using G*Power (*version: 3.1.9.7*, RRID: SCR_013726, Germany) (Faul et al. 2007). The power calculation was made based on the statistical procedures that would be implemented. A moderate effect size (d = 0.5), an alpha error of α = 0.05, and a power of 1-ß = 0.8 indicated that a minimum sample size of 21 participants was required.

### Materials

The data collection took place on the basketball court in the hall of the Institute for Sports Science Center of the University of Oldenburg and once on the training facilities of the EWE Baskets Oldenburg.

We used two camera tripods (CT-10 SmallRig, Shenzhen, China) with two Android smartphones. One smartphone (Samsung Galaxy S21 FE 5G, model SM-G990B2/DS, Android *version 14*, Suwon, South Korea) was used for wireless mobile EEG and video recording with the Smarting Pro application from mBrainTrain, Belgrade, Serbia (*version 3.2*). The second phone (Samsung Galaxy S21 FE 5G, model SM-G990B2/DS, Android *version 14*, Suwon, South Korea) was used for the simultaneous real-time motion tracking with the Pose Landmark Detection (PLD) (Bazarevsky et al. 2020) from MediaPipe studio (Lugaresi et al. 2019). MediaPipe's embedded PLD uses a series of convolutional neural networks to recognize human pose landmarks in real-time, facilitating full-body human motion tracking. The algorithm determines the position of 33 pose landmarks on 2D video data. These landmarks are described in cartesian coordinates. The basic data unit of these coordinates is a packet, which consists of a value class with their own numeric timestamp. They were recorded through Lab Streaming Layer (LSL; Kothe et al. 2024) at a sampling rate of 15 Hz using the customized standalone Android PLD app *version 1.1* (Haupt et al. 2026). The results of on-device real-time body pose tracking research have been used in fitness tracking (Bazarevsky et al. 2020), attention monitoring (Hossen and Uddin 2023), real-time location of acupuncture point location (Malekroodi et al. 2024), and social interaction (Figari et al. 2024). In this study, we used PLD to record motion in natural settings along with EEG.

EEG data were measured with a head-worn EEG Smarting Pro (mBrainTrain, Beograd, Serbia) system with 32 channels (at the 10–20 sites F3, F4, C3, C4, P3, P4, O1, O2, F7, F8, T7, T8, P7, P8, Fz, Cz, Pz, POz, FC1, FC2, CP1, CP2, FC5, FC6, CP5, CP6, TP9, TP10), sending signals wirelessly via bluetooth 5.0. Reference and ground electrodes for the EEG system were placed at FCz and AFz, respectively. Impedances were checked before the recording, and were kept below 20kΩ using an electrolyte gel (Abralyt HiCl, Easy-cap GmbH, Hersching, Germany). The amplifier was attached to the back of the cap, positioned below electrodes O1 and O2 and secured with an elastic headband. EEG data and video data were recorded by the same smartphone using the SmartingPro app at a sampling rate of 250 Hz. Head movements were additionally recorded with an inertial measurement unit (IMU) integrated into the EEG amplifier, consisting of 3D accelerometer, gyroscope and quaternions. The smartphones were placed on tripods at a distance of at least 2 m from the participant.

Motion sensor signals (accelerometer, gyroscope, and magnetometer) were also recorded by a small, lightweight IMU (Movella Dot, Movella Inc., 2023, model Polar OH1+. Bluetooth 5.0, *version 2023.6.1*, Nevada, USA) attached to the right wrist with a wrist strap. The LSL Senda Android app *version 1.0.7* (Blum et al. 2021; Haupt et al. 2026) received the derived data via Bluetooth and streamed them into the local area network at a sampling rate of 60 Hz.

EEG data, PLD data, and motion sensor IMU signals were synchronized using the Android LSL Recorda app *version 1.0.00.9* (Blum et al. 2021; Haupt et al. 2026). Data were stored in a single Extensible Data Format (XDF) file. The first smartphone independently recorded EEG signals. The second smartphone recorded both PLD and motion sensor signals, while running Senda and Recorda apps in the background on that same device.

## Procedure

Participants silently performed continuous free-throws at their own pace, following the general rules of basketball. The shooters took their place at the free-throw line, which was 4.6 m away from the basket. After EEG preparation, participants had 5 min of familiarization before the task. They took a total of 120 shots in 6 blocks, 20 shots per block, with a 1-min break in between to reduce the risk of fatigue and loss of concentration. Participants were asked to perform free-throw shooting with the following instruction: "Prepare for your shot and take your time, then you can shoot whenever you are ready. Do your best!". Participants were not allowed to dribble or bounce the ball between shots. They were encouraged to prepare for their shot before taking it, so that they were all set up and ready to shoot just before the action. The precise timing of the preparation process was not controlled in order to ensure that the protocol remained as natural as possible. An experimenter handed the basketball to the participant. In total, the recording sessions, including preparation and follow-up, lasted approximately 90 min. With the explicit consent of the participants, PLD video recordings were made from the lateral profile of the dominant hand (right side, as all participants were right-handed) as well as from the back of all participants. A schematic of the setup is shown in Fig. 1, a visualization of the actual recording setup is shown in a supplementary video (Online Resource 1).

## EEG data analysis

The study combined EEG, PLD, and IMU streams with different sampling rates. As a first step, we used the Matlab function *load_xdf()* with automatic dejittering configuration to handle small timing discrepancies. Since the primary focus of the analysis was on the EEG data, to align the PLD and accelerometer data with the EEG timestamps, a linear interpolation method (Matlab function *interp1()*) was used with the EEG timestamps as query points. This method resulted in an effectively upsampled PLD and accelerometer data time series with the same time stamps as the EEG data time series. Despite the initial temporal alignment of the data streams provided by LSL during recording, discrepancies may still remain due to varying sampling rates and device-specific timing. These differences can result in uneven frame counts and misaligned timestamps between data streams. We addressed this challenge within the data preprocessing stage described before using EEG time series as the reference. Thus, this procedure ensured precise temporal alignment with the EEG data without altering the inherent properties of the other signals.

## Onset detection

To analyze the RP, we first identified a prominent marker of single basketball shots using PLD signals. Specifically, the raising of the right hand above the eye level was used, which is referred to as the "set-point" when players are engaged in a shot (Brancazio 1981; Penner 2021). Therefore, we identified the intersection of the PLD right eye and right wrist y-coordinates. This artificial onset did not indicate the onset of the movement but rather served as a time reference for each individual shot. The set-point detection is illustrated in Supplementary Fig. 1. From the set-point, the onset of movement was determined backwards in time by a trial-by-trial reverse computation procedure. To do this, we first determined the mean magnitude of the acceleration signal in the baseline period (from − 2500 ms to − 2000 ms), during which participants stood steadily without significant movement. A threshold for movement onset was then defined as the baseline mean plus one standard deviation (*SD*). The algorithm stepped through the signal sample by sample, indicating movement onset as the last sample below the threshold (Verbaarschot et al. 2015). The corresponding time point was then used to epoch the EEG data and was defined as 0 ms latency or movement onset. A visualization of this method is shown in Supplementary Fig. 2.

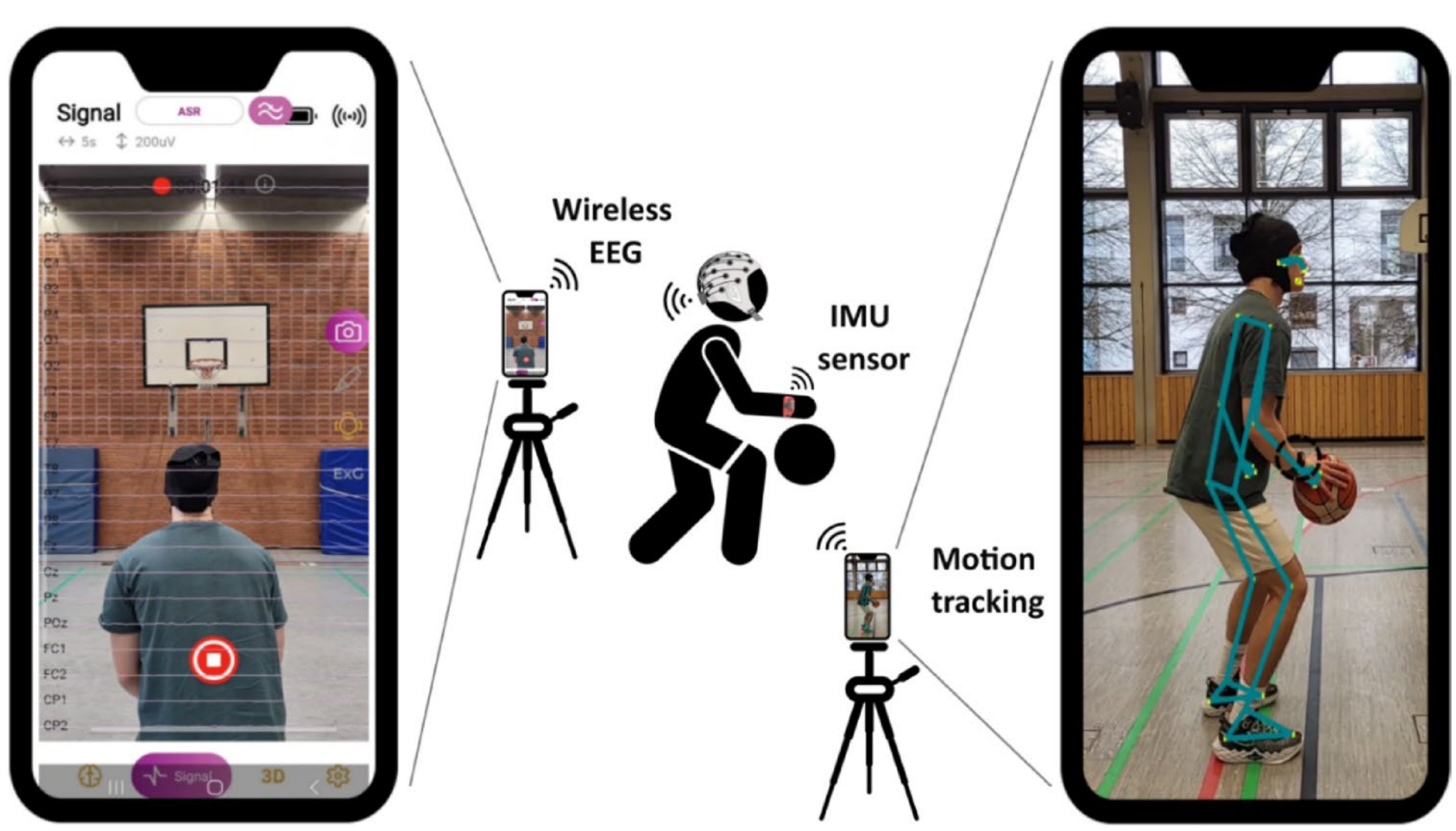

**Fig. 1** Pocketable setup for basketball free-throw shooting. Two tripods are used to keep two Android smartphones in fixed positions. One smartphone wirelessly receives EEG data recorded along with video recordings from the same phone (Smarting Pro app). The second smartphone captures human motion in real-time (MediaPipe Pose Landmark Detection app). A single Movella DOT sensor placed at the right wrist streams IMU signals. LSL Senda and LSL Recorda Android apps manage time-synchronous acquisition of all sensor streams

## EEG preprocessing

EEG data were processed using EEGLAB (*version 2023.0*) (Delorme and Makeig 2004) and custom Matlab scripts (The MathWorks, Inc. 2023; Natick, MA, USA, *version 9.9.0 2023b*). An overview of the processing steps is provided in Supplementary Fig. 3.

First, channels exceeding a root mean square (*RMS*) (Kenney and Keeping 1962) of more than one *SD* above the channel mean *RMS* were flagged as bad and rejected. The data were high-pass filtered at 2 Hz (order 415) and low-pass filtered at 30 Hz (order 111) using zero-phase finite impulse response (FIR) filters (*pop eegfiltnew()*). The data were then epoched from − 2.5 to 0 s (from baseline onset to movement onset) for artifact attenuation. Epochs with strong artifacts were identified using a probability function (*pop_jointprob()* with *SD* = 5). An extended infomax independent component analysis (ICA) was then applied, and the resulting ICA weights were applied to the unfiltered, continuous raw data.

Second, continuous raw data were filtered at 0.2 Hz (order 4127) and 10 Hz (order 331) using FIR filters (*pop eegfiltnew()*). The filtered data were epoched from − 2.5 to 1 s relative to movement onset and baseline corrected (from − 2.5 to − 2 s). Independent components were automatically labeled as containing artifacts using ICLabel (*version 1.3*) (Pion-Tonachini et al. 2019). Bad components were identified if they exceeded a threshold of 90% for 'heart' and 'line noise' artifacts, and 30% for 'muscle', 'eye', 'channel noise', or 'other' (functions *pop_iclabel()*, *pop_icflag()*, and *pop_viewprops()*). The corresponding component activities were then removed from the data by joint back-projection of all remaining component activities. Remaining artifact epochs were rejected (*pop_jointprob()* with *SD* = 3). Spherical interpolation was then used to replace rejected channel data (*pop_interp()*). Finally, the data were referenced to electrodes TP9 and TP10 (*pop_reref()*) and averaged across trials to obtain the RP. Note that the period after movement onset may have contained residual movement artifacts. The focus of the artifact processing pipeline was to obtain a clean pre-movement period as required for RP evaluation. Detailed results of the data lost during EEG preprocessing can be found in Supplementary Table 2.

## Parameterization and analysis of ERP components

On average, 12% of the recorded epochs were discarded, resulting in a mean of 103 remaining trials per participant for averaging (range between 97 and 113 for each participant). The RP was obtained by averaging across trials and statistically evaluated at (fronto-)central channels Cz, C3, C4, Fz, FC1, and FC2 in the pre-movement phase from − 1500 to 0 ms.

The RP was parameterized by averaging amplitude values in 100 ms bins from − 1500 to 0 ms. This resulted in 15 features per participant. The time interval after movement onset (from 0 to 1000 ms) was included for visualization purposes only and was not included in the statistical analysis.

## Statistical analysis

Statistical analyses were performed using custom Matlab scripts. The alpha level of significance was set at 0.05 for all statistical tests. The Shapiro–Wilk test assessed the normality of the ERP amplitude distributions in each bin (Shapiro and Wilk 1965). Where appropriate, the False Discovery Rate (FDR) correction was applied to control for multiple comparisons (Benjamini and Hochberg 1995).

### Movement onset validation

To statistically evaluate the wrist-worn accelerometer-based movement onset detection procedure, we assessed whether the 33 body landmarks indicated movement onset at that time (i.e., the acceleration magnitude at 0 ms minus the acceleration magnitude of the previous sample). The magnitude of acceleration of each body part was calculated and evaluated at the single-subject level by performing Wilcoxon signed-rank tests against zero (Wilcoxon 1945). A correction for multiple comparisons was applied across body parts within participants. Binomial tests were used to explore whether a significant number of participants showed an effect (Combrisson and Jerbi, 2015).

### Presence of the RP

The presence of RP across subjects was evaluated using t-tests on preselected channels and for all time bins covering the interval from − 1500 to 0 ms. Correction for multiple comparisons was applied across channels and time intervals. This analysis identified several time bins confirming the RP, and for these time bins we subsequently evaluated condition effects, i.e., whether a difference in RP could be observed for successful versus unsuccessful trials.

#### Relationship between RP amplitudes and performance

To test whether RP differs between successful and unsuccessful free-throw shots at the single-subject level, we evaluated trial-by-trial fluctuations in single-trial amplitudes using point bi-serial correlation. A label vector (hits = 1, misses = 0) was correlated with the single-trial amplitudes. For each time bin and participant, a measure of explained variance was obtained by squaring the resulting correlation coefficients ($R^2$). This analysis was performed on the (fronto-) central channels only and explored whether single-trial amplitudes are systematically different between successful and unsuccessful shots. The False Discovery Rate (FDR) correction method was applied to control for multiple comparisons across time intervals and channels (Benjamini and Hochberg 1995).

#### Relationship between human pose and performance

The point bi-serial correlation analysis approach was also applied to the PLD motion data, again on a single-subject, trial-by-trial level. For each trial per participant, a label vector (hits = 1, misses = 0) was created and correlated with the x, y, and z coordinates of the body landmarks (representing the 33 body parts) at 100 ms intervals. For each time interval and participant, a measure of explained variance was obtained by squaring the resulting correlation coefficients ($R^2$). The time range for this analysis covered the interval from − 2500 ms to 1000 ms. Thus, we examined the relationship between human pose and shooting performance. The False Discovery Rate (FDR) correction method was applied to control for multiple comparisons across time intervals and body landmarks (Benjamini and Hochberg 1995).

## Results

### Movement onset validation

For each participant, Wilcoxon signed-rank tests were used to determine which body parts showed a significant amount of motion at the accelerometer-based movement onset (time 0). A binomial test was used to determine whether the proportion of participants showing significant movement in each body part was significantly above chance level. We found significant movement across participants in the left index, left pinky, left thumb, left wrist, right index, right pinky, right thumb, right wrist ($p < 0.05$*). A visual representation is shown in Fig. 2.

### Presence of the RP

At the group-mean level, there was clear evidence for RP event-related potential, with both the topography and the morphology of the signal being similar to previous reports of the RP. At the individual level, variability in RP morphology and topography was evident. While the majority showed a clean representation of the RP, the influence of residual artifacts cannot be excluded. Across the − 1500 to 0 ms interval at channel Cz, 57.7% of participants (15/26) showed negative mean amplitudes, ranging from − 22.99 to − 1.43 µV ($SE = 0.37$ to 3.19 µV). These negative deflections were statistically significant in all 15 participants ($z = -8.74$

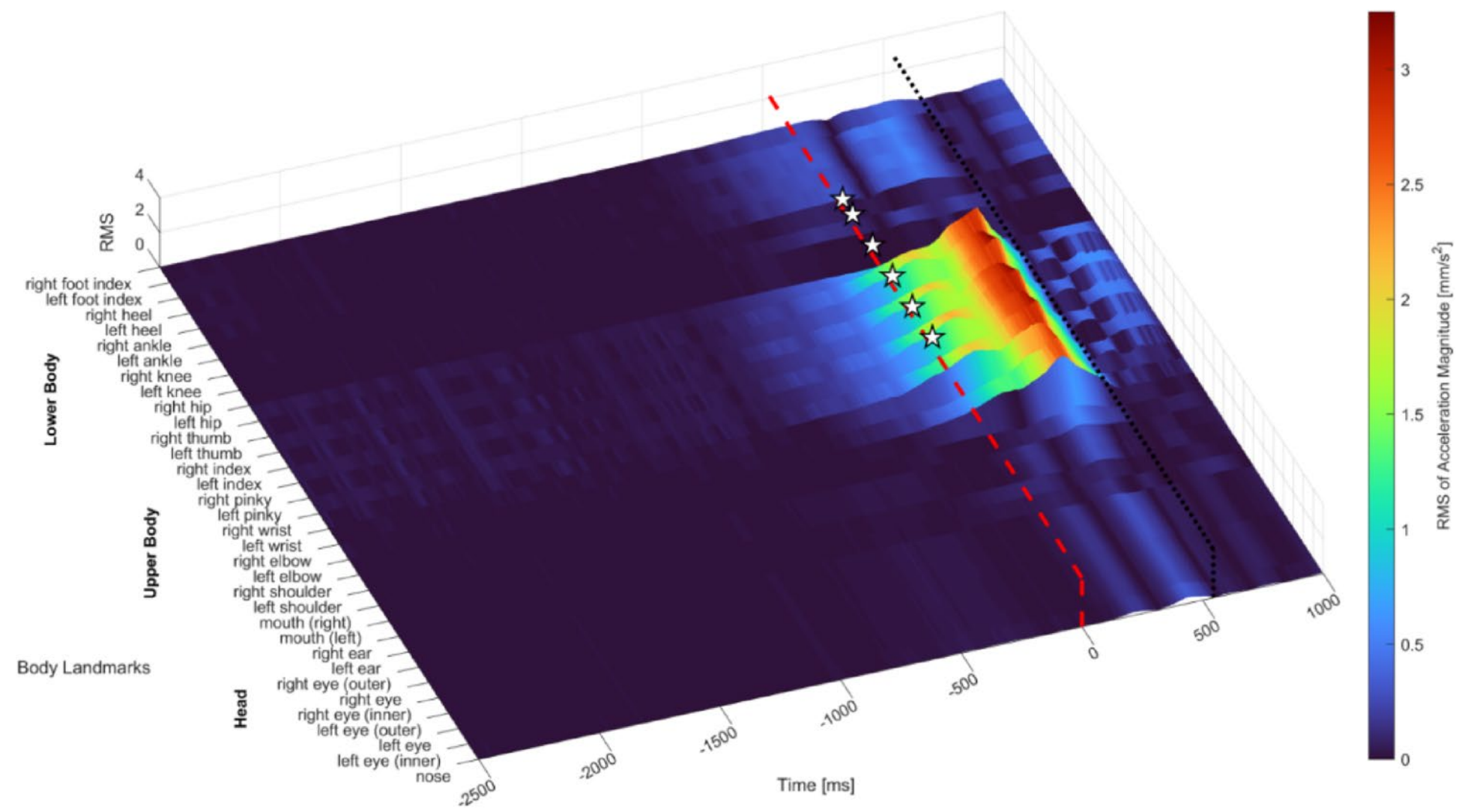

**Fig. 2** Root mean square (RMS) of acceleration magnitude over time across body parts. Results of binomial test for all participants were conducted to identify time intervals and body landmarks where the observed RMS acceleration, computed in 10 ms windows, significantly deviated in relationship with the movement onset. The black dotted line marks the "set-point" (eye-wrist intersection) as a time reference. The vertical dashed line (red) represents the movement onset (time 0). Body parts with significant movement at wrist-accelerometer movement onset are indicated by white stars ($p < 0.05$, FDR corrected)

to − 2.08, $p<0.05$, $d=-0.16$ to − 1.89). In contrast, 42.3% of participants (11/26) showed positive mean amplitudes, ranging from 0.46 to 13.22 µV ($SE=0.58$ to 4.34 µV), with significant positive deviations from zero in 23.1% of participants (6/26; $z=2.57$–8.36, $p<0.05$, $d=0.27$–1.64). Thus, although the group-level RP showed the expected fronto-central negativity, individual Cz waveforms varied considerably in polarity and magnitude, suggesting substantial inter-individual variability in RP morphology or possible residual artifacts. A graphical representation of individual RP (event-related potentials) is shown in Supplementary Fig. 4.

In the grand average RP, the topographic maps provide a clear visual representation of the spatial and temporal distribution of activity, indicating an onset of the RP at approximately − 1000 ms prior movement. The spatial distribution is consistent with the expected (fronto-) central localization of pre-movement neural activity. We tested the grand average RP using the Wilcoxon test for each of the 15 time segments (each segment consisting of the mean amplitude over 100 ms intervals) ranging from − 1500 ms to movement onset. For channel Cz, statistical analysis revealed significant amplitude deviations from zero in four consecutive time segments before movement onset: − 400 to − 300 ms ($M=-6.54$ µV, $SE=\pm2.26$, $z=-2.53$, FDR-corrected $p=0.049$, $r=0.50$), − 300 to − 200 ms ($M=-8.35$ µV, $SE=\pm2.30$, $z=-3.09$, FDR-corrected $p=0.013$, $r=0.61$), − 200 to − 100 ms ($M=-10.54$ µV, $SE=\pm2.25$, $z=-3.59$, FDR-corrected $p=0.004$, $r=0.70$), and − 100 to 0 ms ($M=-13.52$ µV, $SE=\pm2.42$, $z=-3.92$, FDR-corrected $p=0.003$, $r=0.77$). Thus, the Cz RP became progressively more negative as movement onset approached. We also analyzed the other (fronto-) central channels, namely channels C3, C4, FC1, Fz, and FC2. After correcting for multiple comparisons, these channels largely followed the same statistical significance from − 400 ms to − 200 ms ($p<0.05$*) and from − 200 ms to 0 ms ($p<0.01$**), except for channel C4, which was statistically significant only from − 100 ms to 0 ms ($p<0.05$*). Detailed results for individual RPs at channel Cz are shown in Supplementary Table 3.

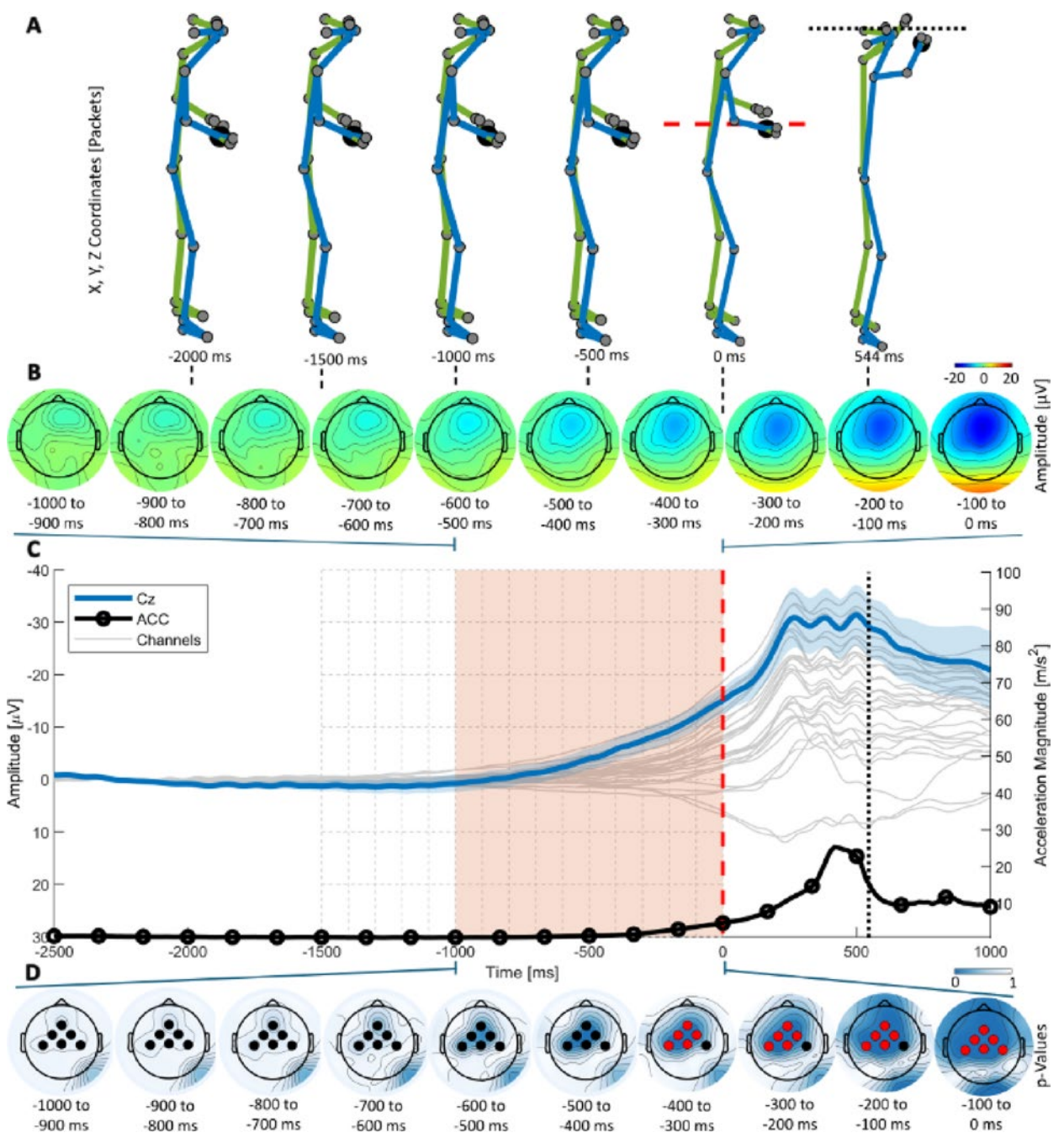

**Fig. 3** Grand average of simultaneous human motion capture and ERP. Mobile EEG, motion patterns, and sensor IMU signals were combined to analyze the readiness potential (RP) and motor activity during task execution. **A** Motion tracking of body postures at key time intervals relative to movement onset. **B** Grand average ERP topographies at time intervals show the spatiotemporal evolution of the RP, with increased negativity over (fronto-) central channel sites leading up to movement onset. **C** RP evolution: The RP (blue line, channel Cz) exhibits a gradual negative deflection preceding movement onset (red dotted line). The onset of movement was determined using wrist accelerometer data (black circle-line). Time reference "set-point" is shown in black dot line. **D** ERP significance testing: Mean ERP amplitudes across 100 ms time bins were tested against zero. Topographical maps display significant regions ($p<0.05$, red dots) after false discovery rate correction, indicating where the RP differed significantly from baseline

The grand average result of simultaneous human motion capture and RP is shown in Fig. 3. A dynamic visualization over time can be found in the animation (Online Resource 2). The $p$-values resulting from the Wilcoxon signed-rank test comparing the grand ERP to zero for all channels are depicted as topographical representations. In addition, human position data are shown at selected intervals. On average at the group level, the accelerometer data showed that the set-point (eye-wrist intersection) occurred 544 ms after movement onset.

## Relationship between RP amplitudes and performance

For each participant, successful and unsuccessful trials were separated and statistically analyzed. For none of the participants and none of the 15 time window bins a significant point bi-serial correlation was found ($p>0.05$). A graphical representation of the comparison of the mean amplitude per bin over the grand mean ERP is shown in Fig. 4.

At the group level, we compared trial-averaged RP for successful and unsuccessful trials and found no significant effects ($p>0.05$). We found that the maximum RP amplitude prior to movement execution was located in channel Fz. Comparison of single-trial amplitudes across participants between conditions (successful vs. unsuccessful shots) of (fronto-) central channels revealed no significant effects ($p>0.05$). In an exploratory manner, the explained variance ($R^2$) of each channel of interest was analyzed across different time segments (from 1500 ms to movement onset) and participants. We focused on identifying significant changes

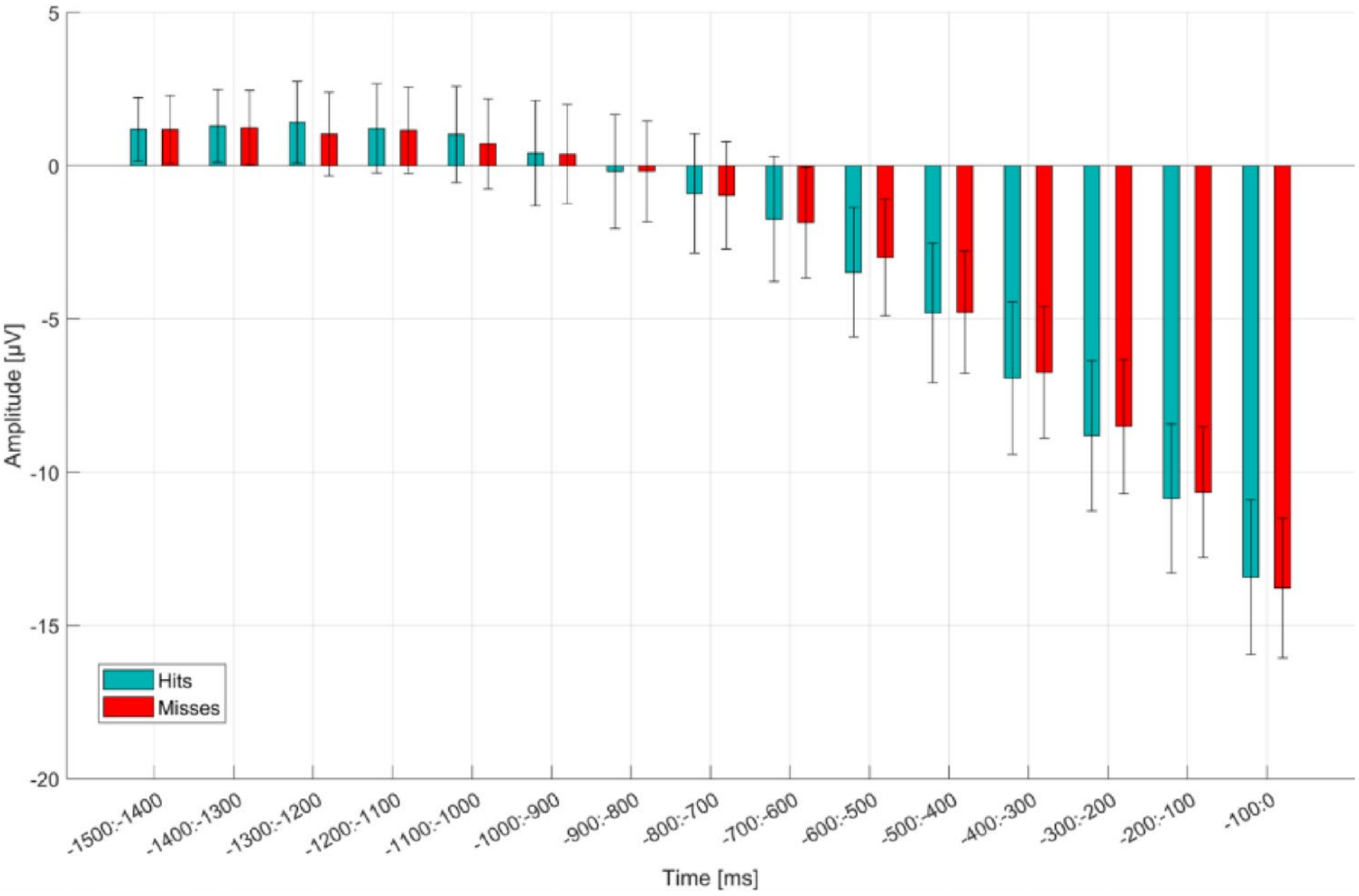

**Fig. 4** Grand average ERP comparison between conditions across participants. The grand average amplitude of the ERP recorded from the Cz channel, comparing successful (hits) and unsuccessful (misses) basketball free-throw shots across all participants is shown. The RP amplitudes are presented in 100 ms bins from − 1500 ms to 0 ms relative to movement onset. Blue bars represent hits, and red bars represent misses, with error bars indicating the standard error of the mean for each bin

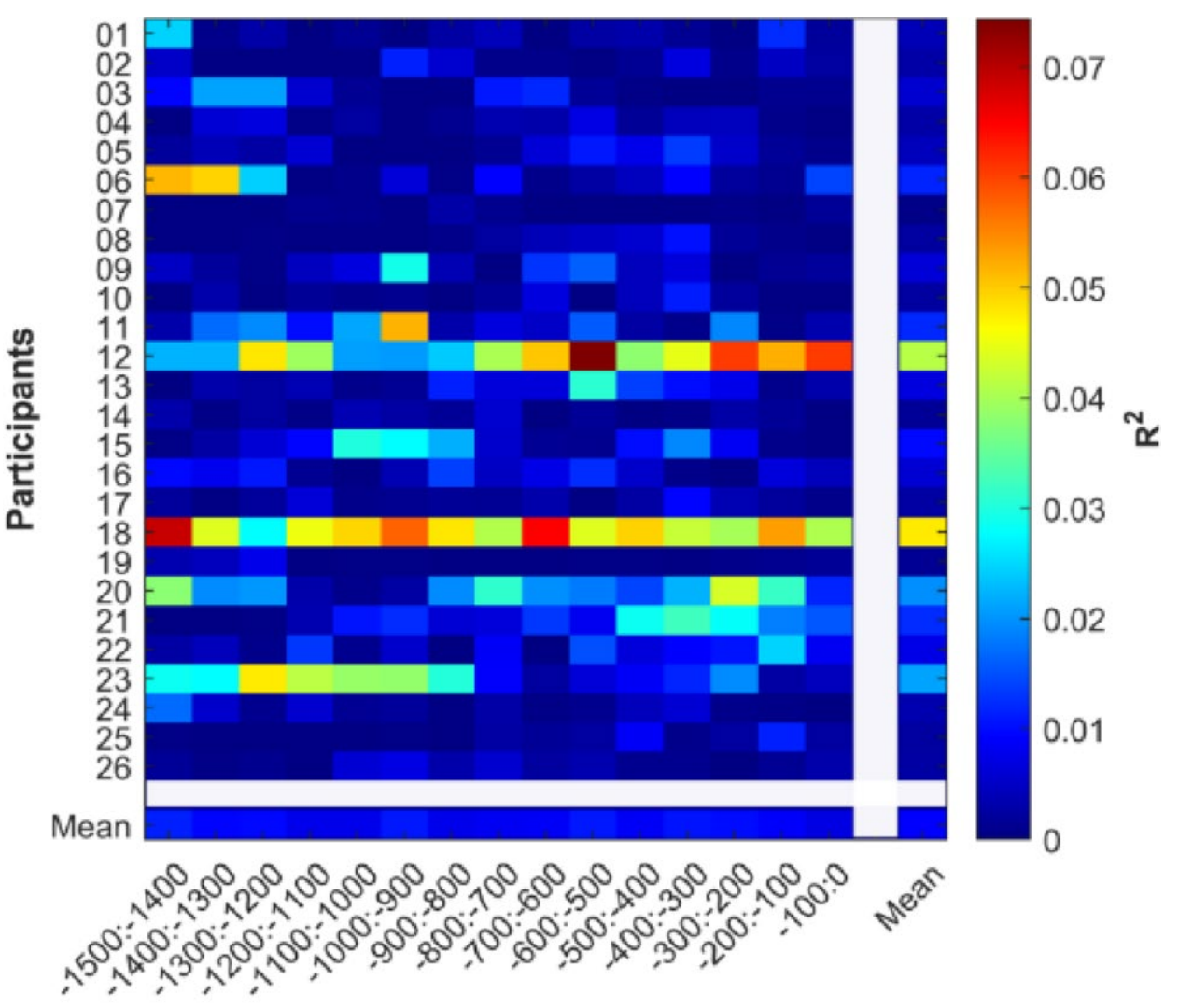

**Fig. 5** Explained variance ($R^2$) from point-bi-serial correlation of single-trial amplitude features between conditions of participants in Fz channel. The heat-map illustrates the $R^2$ of single-trial amplitude features at channel Fz, representing the proportion of variance in trial outcomes (hits versus misses) explained by each feature across 26 participants. Red colors indicate higher $R^2$ values (more variance explained), while blue colors indicate lower $R^2$ values (less variance explained). The time window spans from − 1500 ms to 0 ms relative to movement onset, divided into 100 ms bins. Each row corresponds to a participant, the last column after white space represents the mean $R^2$ values across time bins, and the last row after white space shows the mean $R^2$ values across participants, highlighting the overall most discriminative features for distinguishing hits from misses

in $R^2$ values over time to determine whether basketball shooting performance explained variability in single-trial amplitude features. At the individual level, some participants showed higher differences than others. Specifically, for the Fz channel, participants 12 and 18 had the largest differences between conditions, as indicated by their overall mean $R^2$ over time ($R^2 = 0.041$ and $R^2 = 0.047$, corresponding to 4.1% and 4.7% explained variance, respectively). However, the magnitude of the effect indicated that only a small amount of variance in single-trial RP amplitudes was explained by shooting performance. Furthermore, the overall mean across participants did not exhibit specific differences at any time interval that would indicate performance discrimination. Figure 5 provides an illustration of this analysis for channel Fz, similar illustrations for the other channels of interest can be found in Supplementary Fig. 5.

## Relationship between human pose and performance

In addition to the RP analysis, we explored the difference in motion patterns between successful and unsuccessful shots based on the PLD coordinates obtained from the motion capture software during basketball free-throw shooting.

At the individual level, a point bi-serial correlation was performed to compare the x, y, and z coordinates of each body part between successful and unsuccessful shots. In this case, landmarks were also analyzed at 100 ms intervals from − 2500 ms to 1000 ms relative to movement onset. After FDR correction, the point bi-serial correlation results identified statistically significant differences ($p < 0.05$*) between pose landmarks and shooting outcome within individual

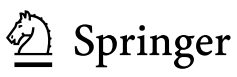

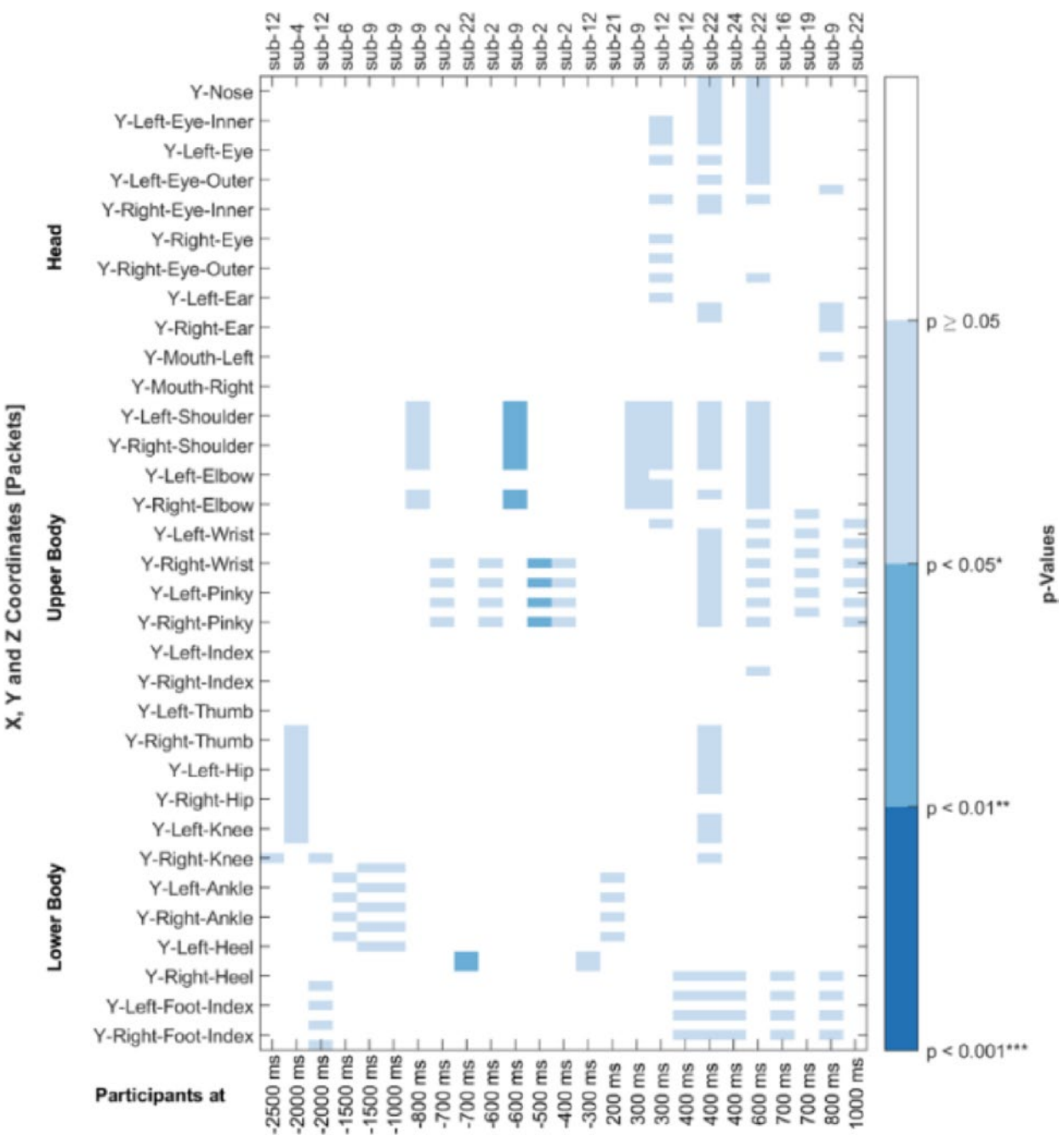

**Fig. 6** Differences in pose landmarks between hits and misses. Point bi-serial correlation analysis identified significant pose differences between hits and misses across participants during specific time windows while basketball shooting. Resulting *p*-values for the X, Y, and Z coordinates of individual body parts derived from motion capture data utilizing Pose Landmark Detection software are shown. Following false discovery rate (FDR) correction, significant differences are color-coded according to their p-value thresholds: $p < 0.05^*$, $p < 0.01^{**}$, $p < 0.001^{***}$. These are concentrated in the upper body landmarks, particularly for the wrists, shoulders, and elbows, suggesting their critical role in differentiating between successful and unsuccessful trials

participants. These findings reflect within-subject differences rather than group-level effects, emphasizing the variability in movement patterns across individuals rather than a generalized trend across people. Particularly, those differences were shown in 38.5% of participants (10/26) (sub-02, sub-04, sub-06, sub-09, sub-12, sub-16, sub-19, sub-21, sub-22, and sub-24) at different time intervals (− 2500 ms, − 2000 ms, − 1500 ms, − 1000 ms, − 800 ms, − 700 ms, − 600 ms, − 500 ms, − 400 ms, − 300 ms, 200 ms, 300 ms, 400 ms, 600 ms, 700 ms, 800 ms, and 1000 ms). Thus, these participants showed significant differences in pose at different time intervals and body parts. However, the amount of variance explained ($R^2$) was variable and low overall.

The *p*-values ($p < 0.001^{***}$) and the overall mean features of body parts over time showed that the upper body had the largest variance of the difference in pose between conditions (successful vs. unsuccessful shots), especially at the hand coordinates "Y-right-wrist" (mean $R^2 = 0.045$, corresponding to 4.5% explained variance), "Y-right-pinky" (mean $R^2 = 0.0426$, corresponding to 4.3% explained variance), and "X-left-pinky" (mean $R^2 = 0.044$, corresponding to 4.4% explained variance). In addition, the overall mean features of each participant's pose revealed that the highest participant-level explained variance in pose at specific time intervals was observed for sub-22 at 400 ms (mean $R^2 = 0.040$, corresponding to 4.0% explained variance), 600 ms (mean $R^2 = 0.043$, corresponding to 4.3% explained variance), and 1000 ms (mean $R^2 = 0.037$, corresponding to 3.7% explained variance), indicating differences in whole-body posture. However, the participants with the most pronounced differences ($p < 0.001^{***}$) indicated different body parts, namely the left shoulder, right shoulder, and right elbow for sub-09 at − 600 ms, and the right wrist, right pinkie, and left pinkie for sub-02 at − 500 ms. A visualization of the results is provided in Fig. 6, a plot illustrating this effect as explained variance can be found in Supplementary Fig. 6.

The evaluation of pose landmarks during the execution of the shot, in particular the Y-coordinates (corresponding to the vertical position normalized to the image height), revealed different movement patterns. For the participants with statistically significant differences, successful performance was characterized by a lower wrist body point relative to the midpoint of the hip (Y-right-wrist: − 0.376 packets) compared to unsuccessful performance (Y-right-wrist: − 0.395 packets) prior to movement execution. However, at the set-point, successful shots were characterized by higher wrist elevation (Y-right-wrist: − 0.161 packets) compared to unsuccessful shots (Y-right-wrist: − 0.154 packets). Finally, at the group level, based on the magnitude of wrist acceleration for each participant, the pose landmark data showed that the set-point occurred approximately 536 ms after movement onset for successful shots and 560 ms after movement onset for unsuccessful shots, revealing a difference of approximately 26 ms between conditions.

## Discussion

In this study, we propose a pocketable setup for the simultaneous acquisition of whole-body human pose data and brain-electrical signals. We investigated the RP in preparation for basketball free-throw shooting. While we did not find differences in the RP in preparation for successful versus unsuccessful free-throws, we were able to capture whole-body movement patterns with an unobtrusive setup consisting of two off-the-shelf smartphones. In some participants body positions in preparation for free-throws differed between successful and unsuccessful throws. Overall, the plausibility of the motion capture and the EEG results confirm the feasibility of our low-cost approach.

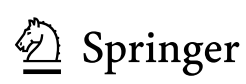

## Movement onset validation

RPs are defined as fronto-central negative deflections in preparation for voluntary actions (Kornhuber and Deecke 1965; Shibasaki and Hallett 2006). It is therefore important to identify the onset of a movement for RP analysis. While this is relatively easy to do for a simple movement such as isolated finger button presses, it is much more challenging for whole-body movement sequences such as jumping or throwing a ball. We used a wrist-worn accelerometer to identify the onset of movement, and whole-body PLD signals to explore its validity. A descriptive evaluation of the PLD signals suggested that hand/wrist motion was among the first, prominent body parts involved in movement onset. Accordingly, we found significant motion of several body parts, such as the left wrist, in the PLD data at the time the accelerometer signal indicated movement onset. While it is beyond the scope of this report to analyze the temporal synchronization of different sensor signals in detail, the overall results suggest a good alignment of EEG, PLD, and accelerometer signals.

As confirmed by binomial tests, the PLD data indicated that movement onset, which we defined by the analysis of wrist accelerometer signals, was evident in upper limb pose landmark signals However, single camera pose detection has limitations. Work on another Media Pipe algorithm called Hand Landmark Detection (Zhang et al. 2020) found that motion capture along the Z-axis (depth) contributed to decreased accuracy of hand pose estimation (Wolke et al. 2025). It is possible that the PLD used here suffers from the same drawback. Accordingly, the camera should be carefully positioned to capture the main motion trajectory of interest. We positioned the PLD camera on the side profile of the participants (vertical axis) and discarded the z-axis pose information for our interpretations. This ensured that the analysis was restricted to the 2D plane.

Computer vision technology for pose estimation is relatively new, and there are models that require large amounts of training data without real-time support, such as OpenPose (Cao et al. 2018), OpenCV (Nour 2021), DeepCut (Insafutdinov et al. 2016), and YOLO (Dong and Du 2024). In contrast, MediaPipe's (Lugaresi et al. 2019) PLD (Bazarevsky et al. 2020) allows for real-time motion tracking. Although its validation for clinical use is still ongoing, early results are encouraging. A recent study has compared different human pose estimation models for motion capture and reported favorable results for PLD (Roggio et al. 2024). Validation studies for clinical applications are also encouraging, suggesting the feasibility of PLD for capturing specific human movement patterns relevant to motor rehabilitation (Latreche et al. 2023; Simoes et al. 2024). The present findings extend to this work by applying smartphone-based

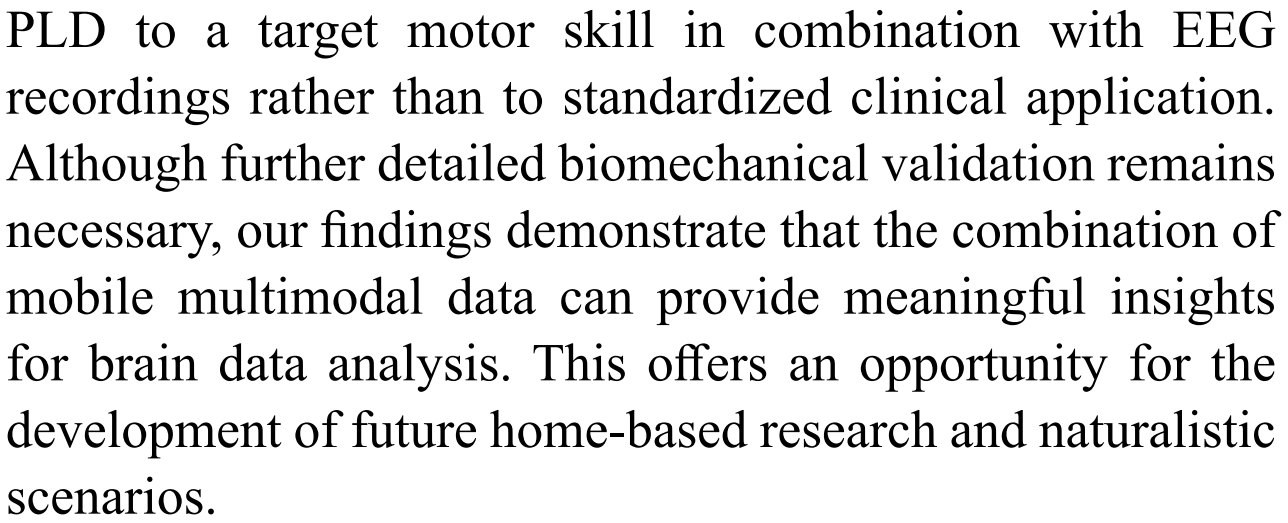

PLD to a target motor skill in combination with EEG recordings rather than to standardized clinical application. Although further detailed biomechanical validation remains necessary, our findings demonstrate that the combination of mobile multimodal data can provide meaningful insights for brain data analysis. This offers an opportunity for the development of future home-based research and naturalistic scenarios.

## Presence of the RP

In the group-averaged event-related potential, a clear RP was evident in the fronto-central channels that preceded the onset of the basketball free-throw movement by several hundred milliseconds. While statistically significant effects were observed from $-400$ ms to movement onset, the RP morphology suggested that the early part of the RP may have begun as early as $-1000$ ms (see Schurger et al. 2021 for a discussion of RP onset and morphology). Previous research on RPs of bilateral motion reported widespread scalp distributions with maximum amplitudes near channel Cz (Deecke et al. 1983; Shibasaki et al. 1980; Shibasaki and Hallett 2006). We found the maximum amplitude at channel Fz, as previously reported by others (Travers et al. 2020). We can only speculate whether this slightly more anterior topography of the RP is somehow functionally related to the subsequent movement pattern. In basketball players, free-throws involve movements of both hands. Thus, a bilateral RP can be expected, in contrast to lateralized RP and movement-related cortical potential topographies for single limb movements (Jacobsen et al. 2022; Tolmacheva et al. 2023). Consistent with previous findings, the slope of the RP increased until movement onset (Kornhuber and Deecke 1965). Basketball free-throw requires considerable movement of the shoulders, arms, and hands, which may lead to movement artifacts in the EEG. Therefore, we do not interpret EEG signals after movement onset. This distinction is important because it shows that mobile EEG can capture interpretable cortical activity even in complex movements, but depuration of strong motion artifacts is still challenging. Thus, the present results support the use of pre-movement EEG markers, such as the RP, as feasible neural targets in naturalistic whole-body tasks, while also highlighting the need for caution when interpreting EEG activity during and after movement execution.

Segmentation of the RP signal into time bins allowed the detection of significant changes in specific time intervals, as was previously done in an RP neurofeedback study (Schultze-Kraft et al. 2021). This method recognizes that the RP is not a uniform signal, but one that evolves over time with different spatiotemporal dynamics between individuals. Overall, the results of both statistical and topographic analyses suggest that the RP is a valid observable phenomenon in this context. Our findings replicate and complement a previous report observing the RP outside the laboratory in preparation for a whole-body movement (Nann et al. 2019).

Given that impedance was not continuously monitored during the recording, gradual changes in electrode contact due to movement, sweating, or cap displacement may have contributed to variability in EEG signal quality and individual RP differences. To minimize these effects, the cap and amplifier were secured with an elastic band, and the preprocessing pipeline included channel rejection, ICA-based artifact attenuation, interpolation, and epoch rejection, based on previous work in mobile EEG (Debener et al. 2012; Jacobsen et al. 2022; Straetmans et al. 2022). In general, the signal quality remained high, only 12% of the data were rejected. However, future mobile EEG studies should consider repeated or continuous impedance monitoring during recording. Additionally, the use of other pre-processing techniques such as Artifact Subspace Reconstruction (Blum et al. 2019; Kim et al. 2025; Kothe & Makeig 2013; Miyakoshi et al. 2023), and Generalized Eigenvalue De-Artifacting Instrument (Ros et al. 2025) may benefit in the preprocessing of highly contaminated mobile EEG data. However, their calibration procedures and parameter settings should be validated to limit excessive correction or removal of neural activity.

## Relationship between RP amplitudes and performance

Previous research suggests that larger RP amplitudes over central cortical areas characterize greater movement preparation and cerebral efficiency (Chiarenza et al. 1990; Crews and Landers 1993; Landers et al. 1994; Mann et al. 2011, Papakostopoulos 1978; Taylor 1978; Konttinen and Lyytinen 1993, Konttinen et al. 1995; Sanchez-Lopez et al. 2014). Thus, larger RP amplitude may precede better movement execution. Evidence for this explanation has previously been found in novice-expert comparisons of target motor skills (Mann et al. 2011; Vogt et al. 2017). However, we did not find an association between RP and free-throw shooting performance. The absence of a similar effect in the present study may be related to the particular shooting task or sample characteristics. Moreover, different mechanisms may be at play when making within-subject versus between-subject comparisons. It is likely that other movement preparation and execution factors, not captured by the RP, play an important role in motor performance.

Previous research has found that significant brain changes related to skill performance are only observed in experienced players (Hatfield and Kerick 2012). Our sample consisted of amateur players with a wide range of experience, ranging from 3 to 18 years of basketball playing. This is reflected in the individual differences in shooting performance, which ranged from 18.1% to 82.6% successful shots. A proportion of our participants may have lacked the training that may be required to initiate specific adaptations leading to improved motor skill performance (Bakker et al. 2021; Hatfield and Kerick 2012; Karni et al. 1998; Mann et al. 2011; Schurger et al. 2021). Given the heterogeneity in skill level exhibited by our participants, a more homogeneous sample, characterized by a higher degree of expertise, may have been more appropriate for the identification of the hypothesized effect. In addition, the typically large interindividual variability in ERP components (Luck 2014; Woodman 2010) may have masked possible associations between RP amplitude and motor performance.

To ensure that we did not overlook participant-specific patterns, we performed a point bi-serial correlation analysis between individual single-trial amplitudes and performance, but this approach also revealed no systematic association between single-trial amplitudes and motor performance. It remains unclear whether the absence of a correlation reflects the absence of an effect or whether it was masked by differences in motor preparation strategies or sample characteristics. It is unlikely that our statistical examination of single-trial activity was strongly biased by the different probabilities of successful and unsuccessful free-throws. At the group level, both classes were almost perfectly balanced, as the average shooting accuracy was 51.9%.

Finally, most previous studies have examined the relationship between RP amplitudes and motor performance in highly controlled environments. The extent to which movement performance can be related to brain electrical activity in natural environments remains underexplored. Initial findings, including the relationship between ERPs and slacklining (Papin et al. 2024), suggest that mobile EEG has considerable potential for detecting individual differences in motor performance. Consequently, the development of accurate assessments of brain-behavior relations in natural scenarios would be valuable in applied areas investigating brain mechanisms in exercise-sports science and neuroergonomics (see Parasuraman 2003; Dehais and Ayaz 2019).

## Relationship between human pose and performance

The movement analysis of some participants revealed significant differences in their posture during movement execution between successful and unsuccessful shots. Successful performance at the individual level was characterized by a lower position of the wrist level in relation to the hip, a higher position of the upper body (arms and hands), and a stable position of the head before releasing the ball.

These results are consistent with previous studies describing proper free-throw technique, from the positioning of the lower limbs, trunk, and upper extremities to the "dip" of the shot (lower ball position relative to the hip), lower knee position, and center of mass. These movement patterns have been associated with greater basketball shooting accuracy (Cabarkapa et al. 2023; Çetin and Murati, 2014; Čoh and Podmenik 2017; Kaya et al. 2012; Penner 2021; Stankovic et al. 2006; Tang and Shung 2005) because basketball players need to lower their bodies to gain strength before motor execution.

While analysis of the full PLD data was beyond the scope of this study, the results of our preliminary analysis are consistent with the conclusion that the PLD captures human pose information reasonably well. At the set-point after movement onset, successful shots were characterized by higher wrist elevation for some participants. This is consistent with previous findings suggesting that the angle, height, and temporal characteristics of the basketball movement are critical for the optimal trajectory of a successful basketball shot (Hamilton and Reinschmidt 1997; Zhao et al. 2022; Tang and Shung 2005; Okubo and Hubbard 2006). This may indicate that high wrist elevation at the set-point is beneficial for a successful outcome. In addition, this stereotypical basketball shooting posture creates a 90° angle between the elbow and wrist angles, representing a "catapult" position to create a curved shot trajectory that increases the probability of hitting the target (Hamilton and Reinschmidt 1997).

For other participants, results also indicated differences in head landmarks between successful and unsuccessful shots, particularly during shot execution. This is consistent with previous research suggesting a relationship between aiming point and accuracy (Gou et al. 2022). In this study, head orientation, which may be indicative of visual attention, could potentially influence shooting accuracy. A stable focus on a specific target point may result in improved performance, whereas an unstable head pattern with shifting visual fixation may have a detrimental effect. This interpretation is consistent with the quiet eye phenomenon, which explains expert-related differences in fixation (Vickers 1996, 2016; Vickers and Williams 2017). Unfortunately, eye tracking data were not available in our study.

The results presented here are case-specific. This suggests that the accuracy of performance is strongly influenced by individual characteristics (Gablonsky et al., 2005; Schmidt 2012; Slegers et al. 2021; Worobel 2020). The case-specific nature of the pose findings has important implications for future research. Rather than assuming that a single group-level kinematic pattern explains successful performance across participants, future studies may benefit from individualized analyses that account for skill level, shooting technique, movement strategy and changes across time. Larger samples, objective expertise criteria, repeated sessions, and predefined biomechanical features such as joint angles, movement velocity, and landmark coordinates alignment may help determine which movement characteristics generalize across participants and those that are case-specific.

## Simultaneous human motion capture and EEG acquisition

We present a novel, easy-to-use, and low-cost setup for capturing the neural correlates of complex, naturalistic whole-body movements. This was achieved by combining camera-based motion capture, an accelerometer sensor, and wirelessly recorded EEG signals. We believe that this new setup can help facilitate brain research in everyday scenarios (Bleichner and Debener 2017). Importantly, the setup is characterized by a relatively high degree of participant and device mobility (Bateson et al. 2017). In other words, it imposes only marginal constraints on participants: they were able to perform basketball throws in a highly natural, ecologically valid manner. According to ECOVAL, a self-assessment tool for ecological validity of experimental setups (Chang, et al. 2022), our experimental setup is very favorable (level 3, score 7–10), equivalent to "natural" and "complex" everyday situations. The use of PLD as a tool for human pose and motion assessment opens the possibility of combining this real-time technology with real-time brain activity measures, which may be useful for brain-computer interfaces and neurofeedback applications.

### Limitations

The high inter-subject variability in skill may have led to human pose distortions in the data in few cases. Moreover, the large variance in event-related potential amplitudes during the basketball shooting implies that some datasets contained a considerable amount of residual noise. However, it was not our intention to describe brain dynamics during the throw and therefore we tailored the processing pipeline towards artifact attenuation of the pre-movement RP interval. The extent to which EEG signals can be evaluated during relatively explosive whole-body movements remains to be determined.

Accelerometer signals were used to identify the movement onset and to validate the temporal relationship between movement and EEG. IMU signals were not included into the EEG artifact processing, based on our previous mobile EEG studies (e.g., Jacobsen et al. 2022; Klapprott and Debener 2024; Straetmans et al. 2024). Although motion signals could provide a rich source of information to correlate with EEG artifacts, sensor artifact modeling was beyond the scope of this study and would require careful validation to avoid removing movement-related neural activity of interest.

We only included participants who were familiar with basketball based on their subjective reports (i.e., at least 3 years of basketball experience). Nevertheless, shooting performance varied considerably among the participants, indicating different skill levels. The use of objective performance measures and inclusion criteria may be relevant for future studies that systematically assess performance within and between individuals. Instruments, such as the wearable occlusion device proposed by Maglott and B. Shull (2019), may help assess cognitive biases in basketball shooting performance, although improvements would benefit the assessment.

Several variables, including minimal delays in synchronization (Iwama et al. 2024), sample characteristics (Bakker et al. 2021; Hatfield and Kerick 2012; Karni et al. 1998), and methodological details such as movement complexity and task demands (Cui et al. 2000; Di Russo et al. 2017; Olsen et al. 2021), movement-onset definition (Russo et al. 2019; Verbaarschot et al. 2015), baseline selection (Alday 2019), filtering parameters and referencing (Hu et al. 2017), and ICA/component rejection criteria (Karimi et al. 2017) may affect RP measurements. In addition, other factors can modulate RP, such as respiration (Park et al. 2020) and concurrent cognitive load, such as mental calculations (Raś et al. 2019). Most importantly, it can be questioned whether the trial-averaged morphology of the RP is a good estimate of single-trial pre-movement activity patterns (Schurger et al. 2021). Our single-trial analyses implied the validity of the additive ERP model (e.g., Makeig et al. 2004), but only in very few participants were we able to find evidence for a continuous evolution of pre-movement activity at the single-trial level, consistent with alternative accounts of the RP (Schurger et al. 2021).

The focus of this work was to explore the feasibility of combining human motion capture and EEG recording signals with smartphone technology during a naturalistic target skill movement. We therefore related pose landmarks to overall performance only in an exploratory manner and did not implement a detailed biomechanical analysis. Such an analysis would require additional methodological constraints and validation, including calibrated camera geometry, more precise 3D reconstruction, joint-angle estimation, ball-release parameters, and possibly multi-camera or marker-based reference data. The present single-camera PLD setup was optimized for portability and real-time feasibility rather than for full biomechanical modeling. Future studies should build on this approach by combining mobile EEG with more detailed kinematic features, such as joint angles, movement velocity, exact release timing, and ball trajectory, to directly test how neural preparation relates to movement execution and performance. Future work may benefit from correlating wireless EEG signals to simultaneously capture human pose and movement patterns in other different contexts.

### Outlook

Future studies may use the combined acquisition of human pose and brain-electrical signals to address basic and applied research questions. While the field of mobile brain-body imaging is growing, the focus so far has been on assessment in laboratory environments. With a sparse, portable and light-weight setup it should be possible in the future to bring recording devices to target individuals and real-life recording conditions.

## Conclusion

To the best of our knowledge, this is the first study to investigate the feasibility of detecting RP during basketball free-throw shooting. This is also one of the first reports of monitoring the RP outside of the laboratory, during the preparation of a complex, whole-body target skill movement. We also conclude that it is feasible to combine human motion capture and human EEG acquisition with versatile, portable, and low-cost technology. This approach, once fully developed and validated, will help researchers to better understand how the brain orchestrates complex whole-body movements. This should be relevant not only in the context of human sports performance, but also for clinical diagnosis and intervention studies.

**Supplementary Information** The online version contains supplementary material available at https://doi.org/10.1007/s00221-026-07342-6.

**Acknowledgements** We would like to thank the professional basketball club EWE Baskets Oldenburg for their cooperation and the Institute of Sports Science at the University of Oldenburg for making the data collection possible. We would also like to thank Reiner Emkes for his technical support.

**Author contributions** MCA: Conceptualization, Data acquisition and curation, Formal analysis, Investigation, Methodology, Project administration, Visualization, Writing—original draft, Writing—review & editing. MK: Conceptualization, Methodology, Supervision, Writing—original draft, Writing—review & editing. NJ: Conceptualization, Supervision, Writing—original draft, Writing—review & editing. PM: Conceptualization, Software, Supervision, Writing—review & editing. JW: Conceptualization, Supervision, Writing—original draft, Writing—review & editing. SD: Conceptualization, Funding acquisition, Methodology, Resources, Supervision, Writing—original draft, Writing—review & editing. MCA wrote the main manuscript text, prepared all figures, and supplementary material. All authors reviewed the manuscript.

**Funding** Open Access funding enabled and organized by Projekt DEAL. The development of software tools used in this study was funded by the Deutsche Forschungsgemeinschaft (DFG, German Research Foundation) as part of the German Excellence Strategy—EXC 2177/1—Project ID 390895285. In addition, internal funds of the Neuropsychological Laboratory Oldenburg were available.

**Data availability** The data that support the findings of this study are available on request from the corresponding author. The MATLAB code is available on GitHub (https://github.com/micufx/Pocketable-MoBI-Baskts). The smartphone LSL apps Senda and Recorda are available on GitHub (https://neuropsyol.github.io/). All information about the set-up is available at https://juliuswelzel.github.io/eeg_basketball_website/.

## Declarations

**Conflict of interest** The authors declare no competing interests.

**Ethical approval** The study was reviewed and approved by the Ethics Committee of the Carl von Ossietzky University of Oldenburg (Drs. EK/2023/087), Oldenburg, Germany. The studies were conducted in accordance with the local legislation and institutional requirements. Participants gave written informed consent to participate in this study. The human images displayed in this paper are the property of participants who have provided their explicit written consent for publication.